# CALIBRATION OF PARALLEL KINEMATIC MACHINE BASED ON STEWART PLATFORM – A LITERATURE REVIEW

**Sourabh Karmakar**
PhD Student
Mechanical Engineering
Clemson University
Clemson, SC
skarmak@g.clemson.edu

**Apurva Patel**
PhD Student
Mechanical Engineering
Clemson University
Clemson, SC
apurvap@g.clemson.edu

**Cameron J. Turner**
Associate Professor
Mechanical Engineering
Clemson University
Clemson, SC
cturne9@clemson.edu

## ABSTRACT

*Stewart platform-based Parallel Kinematic Machines (PKM) have been extensively studied by researchers due to their inherent finer control characteristics. This has opened its potential deployment opportunities in versatile critical applications like the medical field, engineering machines, space research, electronic chip manufacturing, automobile manufacturing, etc. All these precise, complicated, and repeatable motion applications require micro and nano-scale movement control in 3D space; a 6-DOF PKM can take this challenge smartly. For this, the PKM must be more accurate than the desired application accuracy level and thus proper calibration for a PKM robot is essential. Forward kinematics-based calibration for such hexapod machines becomes unnecessarily complex and inverse kinematics complete this task with much ease. To analyze different techniques, an external instrument-based, constraint-based, and auto or self-calibration-based approaches have been used for calibration. This survey has been done by reviewing these key methodologies, their outcome, and important points related to inverse kinematic-based PKM calibrations in general. It is observed in this study that the researchers focused on improving the accuracy of the platform position and orientation considering the errors contributed by a single source or multiple sources. The error sources considered are mainly structural, in some cases, environmental factors are also considered, however, these calibrations are done under no-load conditions. This study aims to understand the current state of the art in this field and to expand the scope for other researchers in further exploration in a specific area.*

**Keywords:** Calibration, Hexapod, Inverse kinematics, Parallel Kinematic Machine

## 1   INTRODUCTION

In the world of conventional manipulators, there are three types of mechanisms: (i) Serial Kinematic Machine (SKM), (ii) Parallel Kinematic Machine (PKM), and (iii) Hybrid Kinematic Machine [1]. A manipulator consists of a base and an end-effector. They are connected by many linkages. A serial manipulator is a machine tool where the linkages are acting in series. A parallel mechanism is formed by connecting a functional body or end-effector which is normally a platform, to a reference body or base through two or more elements forming a closed-loop linkage [2]. A hybrid machine structure integrates SKM and PKM features. The hybrid structure increases the workspace of the end-effector at the cost of the rigidity of the structure. Parallel Kinematic Machine tools have received a lot of attention due to their high dynamic flexibility, structural rigidity, high accuracy due to the closed kinematic loops, no error accumulating characteristics [3], higher load-to-weight ratio, and uniform load distribution capacity compared to the serial manipulators [4]. The number of connecting elements between the fixed base and movable platform varies between 3 to 6 and the number of connections along with the type of connections usually decides the degrees of freedom (DOF) of the machine.

One such PKM controlled by 6 linkages connected between the fixed base and movable platform with 6-degrees of freedom (DOF) is called Hexapod. The first hexapod was designed by Gough in 1947 to test tires with six actuators acting as the linkages. This system had the structure of an octahedral hexapod [5]. Based on Gough's platform, in 1965 Stewart presented a sophisticated 6-DOF platform for using as a flight simulator [6]. The combined motion of the 6 cylinders gives the platform high precision, high structural stiffness, and high dynamic performance [7]. Recently, PKMs based on the Stewart platform have been used in multiple fields. The potential applications of parallel manipulators include mining machines, walking machines, both terrestrial and space applications including areas such as high-speed manipulation, material handling, motion platforms, machine tools, medical fields, planetary exploration, satellite antennas, haptic devices, vehicle suspensions, variable-geometry trusses, cable-actuated cameras, and telescope positioning & pointing devices [8]. They have been used in the development of high precision machine tools by companies such as Giddings & Lewis, Ingersoll, Hexcel, Geodetic, and others [9,10]. The application horizon expanded from the field of a simulator to automobile manufacturing, inspection, human-robot collaboration, space telescope, medical tool control (by adding hexapod at the end-effector point of a serial manipulator) [11]. To achieve the precision and accuracy needed for these machines to perform at the desired level of operational characteristics, the platform movements must be accurately controlled. To obtain the necessary accuracy, it is essential to know the various errors associated with the machine at the time of its operation and identify and implement ways to compensate for the errors. Calibration of the hexapod machine identifies these error factors and adds the correction values to make the output data reliable and predictable [12].

This paper surveys the calibration methods of hexapod kinematic machines based on Stewart platforms. Efforts have been made to include most of the major articles published after the year 2000 which address calibration techniques following inverse kinematics; however, some publications may have been inadvertently overlooked. The paper has been divided into six main sections. The first section serves as an introduction. The second section reviews the kinematics of hexapods and the main error factors that impact the accuracy of hexapods. Sections 3 reviews the types of calibration processes and the strategies used for a successful calibration. Section 4 is intended to cover major calibration methodologies found in the literatures and their outcomes. Section 5 provides a brief discussion and comparisons of the different points captured in the previous section. Finally, section 6 provides a conclusion to the paper.

## 2  HEXAPOD KINEMATICS & ERROR FACTORS

The hexapod kinematics can be solved in two ways depending on the inputs and outputs to the kinematic problem: *forward kinematics* and *inverse kinematics* [13]. In *forward kinematics*, the pose (position and orientation) and velocity of the end-effector are calculated based on the length and orientations of the six cylinders and joint velocities. It can be expressed according to equation 1:

$$[x, y, z, \alpha, \beta, \gamma]^T = f(q_1, q_2, q_3, \dots \dots, q_n) \quad (1)$$

Where, $x, y, z, \alpha, \beta, \gamma$ are pose of the end-effector and $q_1, q_2, q_3 \dots q_n)$ are link variables.

The opposite case is *inverse* or *implicit kinematics* [14]: the length and orientation of the cylinders and joint velocities are calculated based on the desired position and velocity of the end-effector. The equation for this can be given as follows:

$$q_i = f_i(x, y, z, \alpha, \beta, \gamma) \quad (2)$$

Where, $i = 1 \dots n$ are link numbers.

In both cases, the positional accuracy of the hexapod is dependent on a number of error factors [15]. These factors can be geometric or non-geometric [16]. Depending on the error factors, the calibration process is classified into three levels:

- *Level-1 calibration* considers only the joint errors that play a critical role in the accuracy of the robot under dynamic conditions.
- *Level-2 calibration*, also known as kinematic calibration, takes care of the error of the kinematic parameters.
- *Level-3 calibration*, also called non-kinematic calibration captures the errors of non-geometric or quasi-static parameters [17] such as stiffness, the geometry of the robot structure, and errors caused by temperature variation [18].

In a hexapod platform, the components of the rigid structure like the base, frame, top platform, and other accessories are normally fabricated from metals. As such, the dimensional accuracy of these components has a direct influence on the accuracy of the whole system. The dimensions of the structure are dependent on design tolerances or manufacturing deviations, clearances, joint errors [19], thermal deformations [20,21], and elastic deformations [22]. The leg actuators or struts are connected to the structure with movable joints; there can be an error due to the assembly deviations in form of joint run-out and ball screw deviations. Also, the mechanical joints should have low friction, zero hysteresis, and a minimum backlash [3]. If the struts are operated by hydraulic fluids, this can result in transmission errors and sensor errors [23]. Finally, modern hexapods are controlled through software due to the complex nature of their operations and algorithms. So, an error can propagate due to the error in algorithm, execution performance, and truncations in the numerical data.

Therefore, the accuracy of a hexapod can be expressed broadly by the function described in equation 3.

$$Hexpod\ Accuracy = f \begin{pmatrix} mfg.\ deviation, \\ assembly\ deviation, \\ transmission\ error, \\ elastic\ deformation, \\ thermal\ deformation, \\ sensor\ accuracy, \\ algorithm\ error, \\ truncation\ error \end{pmatrix} \quad (3)$$

These error parameters are further elaborated in Table 1. It should be noted that there are more typical sources of errors [24], other than those mentioned here, contribute to the outcome from the system; however, their influences are dependent on the construction philosophy adopted for that particular system, and the type of operations and level of accuracy expected out of those systems.

**Table 1: Error function illustrations**

| Error source | Dependency | Remarks |
|---|---|---|
| Manufacturing deviation | Component's tolerances | These components act as the basic structure of the machine and any deviation have permanent impacts. This structure bears all the loads generated in the static and dynamic condition of the machine and provides rigidity to the machine. |
| Assembly deviation | 1. Manufacturer specification<br>2. Age of the system<br>3. Amount of usage | Generally, assembly deviations are controllable and minimized by the replacement of old, worn-out parts with new parts. |
| Transmission error | 1. Response Time of the actuation system<br>2. Clearance between the links<br>3. Position of the platform<br>4. Speed of operation & lag parameters<br>5. Hysteresis error | This error depends on the robustness of the system and system configuration. Some default limitations cannot be avoided. |
| Elastic deformation | 1. Material used for structure<br>2. Loads acted upon<br>3. Position of the platform | Dependent on the structural materials used and its response property under load. |
| Thermal deformation | Working temperature | Not much change in temperature is expected in normal working conditions. |
| Sensor accuracy | 1. Manufacturers' specification<br>2. Calibration interval | Modern sensors are most accurate. |
| Algorithm | Coding methods | Can be controlled by efficient software and hardware. |
| Truncation | Numerical values | Can be controlled by implementation techniques and with sophisticated systems. |

## 3 CALIBRATION

The purpose of kinematic calibration is to improve the accuracy of the robot kinematic parameters by accurately calculating their values within its defined workspace. A PKM can be calibrated by three strategies (Figure 1): External calibration, Constrained calibration, and Auto or Self-calibration [25].

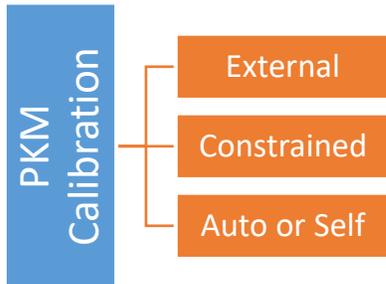

**Figure 1: Strategies for PKM calibration**

External calibration is dependent on the use of external instruments such as an electronic theodolite or a laser tracker [26] for the measurement of multiple poses of the end-effector. In the Constrained calibration process, some motions of the mechanical elements of the robot are constrained to gather the error data. This method is comparatively simplest and least expensive [27]. Auto or Self-calibration is one of the most expensive and complex calibration techniques. In this method, the robot itself automatically takes care of the error parameters measured by redundant sensors with the help of the built-in algorithms installed on controllers. The error correction process can take place during the normal robot operation. Several extra sensors are installed in the joints and linkages of the robot to gather calibration data continuously. Generally, the number of sensors used in a parallel machine is equal to its number of degrees of freedom. It should be noted that for all these strategies, sensors play an important role and they are an essential part of the calibration process; the difference occurs on how these sensors are employed.

Conventionally, the four steps shown in Figure 2 are followed for the calibration process of a PKM: kinematic modeling, measurement, identification, and implementation [28–30].

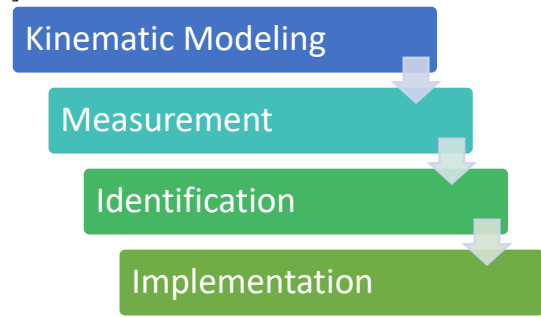

**Figure 2: Steps in PKM calibration process**

Kinematic modeling of the platform is to build the relationship between the joint variables and the platform pose, along with the measurement device readings. The result of the modeling phase is a set of analytical equations showing relationships between these parameters [31]. The measurement phase gathers the data related to the actual platform position and orientation with the help of the measurement devices [32]. The identification step will identify the optimal set of unknown parameters based on the kinematic model and measurements to fit the actual behavior of the mechanism considering the major sources of errors for the context [7,33] and finally, through implementation, the compensation for the calculated model errors are added in the manipulator controller [30].

The results of a robot calibration process are expressed in terms of the position and orientation error values for a set of locations and orientations [34]. To generate a reliable accurate result, the mechanical structure of the robot must be defined with an adequate number of parameters without repetition in the calibration model. A "good" calibration model must have three criteria: completeness, equivalence, and proportionality. Completeness refers to the fact that the model must have enough parameters to completely define the motion of the robot. Equivalence means that the derived functional model can be related to any other acceptable model. And the proportionality property must give the model the ability to reflect small changes in the robot geometry with small changes in the model parameters [35]. A "good" calibration model can be established by building relationships with independent parameters that are used to define the robot system. For a multiloop parallel robot such as the 6-DOF hexapod, the number of independent parameters (C) can be calculated as follows [36]:

$$C = 3R + P + SS + E + 6L + 6(F - 1) \quad (4)$$

where…
- L, number of loops in the robot,
- R, number of unsensed* revolute joints
- P, number of prismatic joints
- S, number of unsensed* spherical joints
- SS, number of pairs of S-joints connected by a simple rod without any intermediate joint
- E, number of measurement devices or transducers
- F, number of arbitrarily located frames

* joint is not equipped with a sensor.

For a Stewart platform with universal (U), prismatic (P) and spherical (S) joints for each leg considering [UPS ≈ 2RP3R], the minimum total number of independent parameters necessary for the complete calibration model is as follows.

$$C = 3 * 6 * 5 + 6 + 0 + 6 + 6 * 5 + 6(2 - 1) = 138$$

## 4 CALIBRATION APPROACH

To achieve a high level of kinematic accuracy, it is necessary to formulate a robust and reliable calibration method. After the introduction of hexapod machines by Stewart, calibration became the field of interest for several researchers. This research is done in various ways: through simulation and physical experimentations. The simulations are not real calibration methods, and they cannot be categorized under any of the three calibration methods discussed earlier, still, they are worthy to study to get an idea of the research direction on the hexapod calibration process.

### 4.1 Simulations for Calibrations

Recent use of dual quaternions by the researchers for the equations of motions resulted in superior performance improvements in terms of computational efficiency while controlling 6-DOF hexapod robots by inverse kinematics [37]. Though this research does not focus on any specific error parameter for the robot accuracy improvement, it has shown improvements in computational efficiency of 43.45% & 38.45% for hexapods consisting of 6-UPS & 6-PUS respectively over the traditional approach.

Agheli et al. [38] considered the lengths of the actuators as the main sources of error for the hexapod accuracy and accordingly modeled their calibration processes to compensate for the platform position and orientation errors. For his simulations, Agheli et al. used a least square method based on the Levenberg-Marquardt algorithm for optimization.

In [14], the researchers Daney et al. presented a method of calibration based on interval analysis. However, their simulation with this method did not satisfy all possible solutions and failed to generate real results. In another study, Daney et al. [23] used the constrained optimization method and DETMAX for inverse Kinematics applications. In this case, they studied the error impact of joint position and leg lengths. They found the mean error on kinematic parameters improved from 0.2705 cm to 0.0335 cm for random poses and 0.2705 cm to 0.0023 cm for selected poses.

Daney et al. carried out two more simulation studies. In one study [39] using algebraic variable elimination and monomial linearization yielded superior results with small numbers (approx. 10) of measurements for their hexapod calibration. In the next one [40], they used Algebraic Elimination to calibrate the platform pose with the help of the kinematic parameters. They got an initial pose error reduction by 99% and 80-98% respectively by the above two elimination methods.

Wang et al. [41] discovered that the errors from the actuator lengths are the most dominant error factor for the overall accuracy of the 6-DOF platform. This analysis considered the errors from the actuator lengths, ball joint location and motion errors.

A summary of research on simulation-based calibration is shown in Table 2. Note that all cases reviewed use inverse kinematics.

### 4.2 External Approach

Discussion of external approaches to calibration has been summarized in Table 3 and Table 4. The instruments used, type of error considered, and kinematics are presented in Table 3; whereas the methods, performance, and key findings of the study are presented in Table 4.

**Table 2: Summary of simulation work on hexapod calibration**

| Reference | Types of error considered | Methods | Performance | Important points |
|---|---|---|---|---|
| Yang et al. 2019 [37] | Platform pose | 1. Dual Quaternions. 2. Virtual power to construct the EOM (equations of motion) | Computational efficiency increased 43.45% for 6-UPS & 38.45% for 6-PUS over traditional approach. | 1. Concise parameterization of translations, and rotations 2. Avoid singularity in motion control |
| Agheli et al. 2009 [38] | Actuator length | A least-square method based on Levenberg-Marquardt algorithm | The position and orientation error reduced 500 and 15000 times, respectively. | - |
| Daney et al. 2004 [14] | Leg length | Interval method | - | The least square method does not satisfy all possible solutions. |
| Daney D. 2002 [23] | Joint position and leg length | Constrained optimization method and DETMAX | Mean error on kinematic parameters improved from 0.2705 cm to 0.0335 cm for random poses & to 0.0023 cm for selected poses. | The error value decreases steadily with an increase in the number of randomly chosen poses and remains usually constant for carefully chosen configurations. |
| Daney et al. 2001 [39] | Platform pose | Algebraic Elimination | Initial pose error reduced by 99%. | 1. An efficient technique to enhance the robustness of the measurement process. 1. Possibility of this method for the self-calibration process. |
| Daney et al. 2004 [40] | Leg length | Algebraic variable elimination and monomial linearization | Initial pose error reduced by 80-98%. | 2. No hypothesis for noise distribution 3. A superior method for small numbers of measurements |
| Wang et al. 2002 [41] | Actuator length, location/motion error of ball joints | A generalized error model | 1 mm actuator length (z-axis) error causes platform deviation in x-axis from -140 to -180 µm, y-axis from 520 to 650 µm, z-axis from -150 to -350 µm | The length error (z-direction) of the actuators influences the accuracy of the machine much larger than any other errors. |

**Table 3: Summary of external approaches (part 1)**

| Reference | Robot Name | Instruments used | Type of error considered | Kinematics |
|---|---|---|---|---|
| Mahmoodi et al. 2014 [42] | - | 6 rotary sensors on 6 legs | Platform pose | Forward/Inverse |
| Jáuregui et al. 2013 [15] | - | Laser interferometer | Actuator length | Inverse |
| Ren et al. 2013 [27] | XJ-HEXA | Biaxial Inclinometer | Actuator length | Inverse |
| Nategh et al. 2009 [43] | Hexapod table | Image capture system | Platform pose | Forward/Inverse |
| Großmann et al. 2008 [44] | FELIX | Double Ball Bar | Model-based factors for thermal and elastic errors. | Inverse |
| Liu et al. 2007 [45] | - | 3D Laser Tracker | Leg lengths | Inverse |
| Ting et al. 2007 [46] | Micro-positioning platform | DMT22 Dual Sensitivity Systems and C5 probe | Hysteresis of Piezoelectric actuators | Inverse |
| Daney et al. 2006 [47] | DeltaLab's "Table of Stewart" | A 1024-768 CCD camera. | Joint imperfection and backlash of each leg. | Inverse |
| Dallej et al., 2006 [48] | - | Omni-directional camera | Position and orientation of the legs | Inverse |
| Daney et al. 2005 [49] | DeltaLab's "Table of Stewart" | Sony digital video camera | Joint position and leg length | Inverse |
| Gao et al. 2003 [50] | FFCM of FAST | Laser Tracker LTD500 | Platform pose error, joint position & leg length | Inverse |
| Renaud et al. 2002 [51] | - | A camera & a LCD monitor | Platform pose | Inverse |
| Week et al. 2002 [52] | Ingersoll HOH600 | Double ball bar redundant leg | Leg length, Gravity load & thermal deflection | Inverse |
| Ihara et al. 2000 [53] | - | Telescoping magnetic ball bar (DBB) | Length error of Struts, position error of base & joint errors of platform | Inverse |

**Table 4: Summary of external approaches (part 2)**

| Reference | Methods | Performance | Important points |
|---|---|---|---|
| Mahmoodi et al. 2014 [42] | A new method with rotary sensors | Positional and orientation variances improved to 0.16 m$^2$ & 0.16 rad$^2$ respectively for small & moderate movements. | • The new method is reasonably accurate, simple to implement, more practical, less expensive, and easier to maintain.<br>• The method is less sensitive to the direction of motion and geometry of the platform. |
| Jáuregui et al. 2013 [15] | Simplified method (all actuators have the same amount of errors) & comprehensive method (each actuator error is not equal) | Most of the result values fall within 10μm of accuracy. | • The simplified method creates linear relationships and easy to solve.<br>• The comprehensive method is complex, non-linear, but more accurate. |
| Ren et al. 2013 [27] | Keeping any two attitude angles of the end-effector constant. | • Position accuracy = 0.1mm<br>• Orientation accuracy = 0.011° | • Exempting the need for precise pose measurement and mechanical fixtures.<br>• Independent of inclinometer range and accuracy. |
| Nategh et al. 2009 [43] | A least-square approach based on Levenberg-Marquardt algorithm. | The position & orientation errors as per simulation were 0.01mm and 1" respectively and were 0.1mm and 0.03° as per experiment. | Employed Observability index to find the most visible & optimum number of measurement configurations. |
| Großmann et al. 2008 [44] | Genetic Algorithm based Trajectory optimization. | The deviation of up to 0.7 mm is reduced to 0.17 mm. | Genetic algorithms are slow to get the most accurate solution, also rarely improve the solution. |
| Liu et al. 2007 [45] | Genetic Algorithm | After 5000 generations the platform position & orientation improved 1.4 & 2.4 times respectively without measurement noise filter. | The genetic algorithm showed good calculation stability, though is not sensitive to measurement noises. |
| Ting et al. 2007 [46] | Preisach model | Platform accuracy level achieved 1 μm in position and 10 μ deg in orientation. | The convergence of errors for fixed points can happen after several iterations. |
| Daney et al. 2006 [47] | Interval arithmetic and analysis methods | Yielded intervals for the position and orientation, with noise and robot repeatability error. | Finds ranges of parameters that satisfy the calibration model. |
| Dallej et al., 2006 [48] | Linear regression | Experimental validation of the method yielded approximately 1 cm median error. | • The method can be applied to determine the geometry of the moving platform.<br>• No mechanical modification of the robot and no calibration pattern are necessary. |
| Daney et al. 2005 [49] | Constrained optimization method and Tabu search | Improvements in accuracy were not as per expectation due to the biasness error of 1.29mm on the z-axis. | • The workspace boundary has a concentration of optimal poses.<br>• By increasing the observability index, the robustness of measurement increases. |
| Gao et al. 2003 [50] | Least Square method | Accuracy improved to 0.2mm | • Method is effective even in lack of enough measurements.<br>• Some false parameters may occur for fewer measurement configurations. |
| Renaud et al. 2002 [51] | Error function minimization. | Precision in the order of 1 μm for an axial stroke of 400mm | • Low cost and easy to use.<br>• Difficult to get a high ratio of accuracy / measurement volume. |
| Week et al. 2002 [52] | - | • Roundness accuracy improved by 3.7x<br>• Squareness accuracy improved by 7x. | The redundant leg was able to predict and compensate for the deflections due to gravity. |
| Ihara et al. 2000 [53] | Fourier transformation | Machine's motion error decreased to ¼. | The measurement is easy and the result on a polar chart is easy to analyze. |

Mahmoodi et al. [42] proposed a new method of calibration for Stewart platform-based PKM. The method is not too sensitive to the direction of the movement and geometry of the platform. In this study, 6 rotary sensors were used in 6 legs to correct the pose of the platform. They used a mix of forward and inverse kinematics for their PKM and observed the positional and orientation variances to be 0.16 m$^2$ and 0.16 rad$^2$ respectively for both small and moderate movements.

The studies by Jáuregui et al. [15] used a laser interferometer as the measuring instrument to calibrate the hexapod. They used inverse kinematics and considered the error related to the actuator length. Their experiment consisted of two methods. In the first method they considered the error from all legs to be the same, applying linear relationships. In the second method, labeled it as the comprehensive method, each leg errors were measured separately and modeled in a non-linear relationship. As expected, the second method resulted in complex calculations but yielded greater accuracy. Ren et al. [28] did their experiment with their PKM named XJ-HEXA using a biaxial inclinometer with the repeatability of 0.001° and length precision of 0.002 mm and reached a position and orientation accuracy up to 0.1 mm and 0.01°respectively after calibration at 80 configurations.

Nategh et al. [43] studied their "Hexapod Table" with the use of an image capturing system. In this research, the results obtained by the simulations and experiments matched very closely. The platform position & orientations errors as per the simulation were 0.01mm and 0.00027° respectively, whereas those for the experiment were 0.1mm and 0.03°, respectively. A least-squares approach based on the Levenberg-Marquardt algorithm was employed in this calibration process. In another study by Großmann et al. [44], a Double Ball Bar (DBB) was used to identify and collect the kinematic parameters by moving the platform on a specific trajectory in the 3D workspace. They used a genetic algorithm and simulated measurements to finalize the parameters. Their hexapod named FELIX was designed and manufactured with a focus on simplicity and capacity for compensating the motion errors generated due to the thermal and elastic deformations. The thermal and elastic deformations were considered in the algorithm by incorporating fixed factors. By this method, they were able to reduce the initial deviation of 0.7mm to 0.17mm after optimizing the trajectory orientation through kinematic calibration. Liu et al. [45] also used a genetic algorithm for calibrating their hexapod using inverse kinematics. They measured the errors coming from the leg lengths and used a 3D laser tracker for the measurements. For their experiments, though they obtained an improvement in the position (1.4x) and orientation (2.4x) of the platform, they found that the genetic algorithm is not sensitive to the measurement noises.

The applications of hexapod did not remain restricted to the dimension level of mm or inch, it has attracted attention for micro-level applications too. Ting et al. [46] did their experiment with a micro-positioning platform to evaluate the hysteresis of the piezoelectric actuators. By using inverse kinematics with the Preisach model, they achieved platform accuracy level 1 µm in position and 10 µ deg in orientation after several iterations. For their experiment, they used Lion Precision DMT22 Dual Sensitivity Systems in association with C5 Probe for measurement.

A popular method of PKM calibration includes vision-based data collection. Daney et al. [47] employed calibration processes using external instruments like CCD cameras and Dallej et al. [48] used omnidirectional cameras. In both pieces of research, the error related to the legs had been investigated by using the images from the external cameras. Then the data were utilized for calibration by using methods like linear regression and interval arithmetic methods. By using the omnidirectional camera, Dallej et al. got a median final platform error of 1 cm.

Researchers Daney et al. did several experiments with hexapod and other PKMs before and after the year 2000. Their work [49] on the machine DeltaLab's "Table of Stewart" involved Sony digital video camera for measuring the joint positions and leg lengths. In this case, they used a Constrained optimization method and Tabu search, but the results obtained were not satisfying due to the error resulting from bias of the system along the z-axis in the range of 1.29mm.

Gao et al. [50] carried out their study and calibration of a Five-hundred-meter Aperture Spherical Telescope (FAST) using inverse kinematics. They used a laser tracker for measurements and controlled the position and orientation of the platform with a Stewart platform-based Fine Feed Cabin Model (FFCM). In their study, they were able to achieve the desired accuracy level of 0.2 mm for the FAST. Here they had not considered any specific error factor except the final pose of the telescope.

In their research, Renaud et al. [51] used a camera and an LCD monitor to calibrate a 6-DOF PKM by using inverse kinematics. Here the final pose of the platform was studied to measure the final accuracy. Though the error parameters captured in the process were not specified, from the discussion and result it is understood that the errors of the actuator lengths were considered to find the correction factors for the final pose of the platform. The findings by Week et al. [52] added errors due to leg length, thermal deflections, and gravity load impact on the end-effector in the calibration process of the hexapod. The use of Double Ball Bar (DBB) as a redundant leg in their Ingersoll HOH600 robot was able to predict and compensate for the error of the platform because of gravity acting on the end-effector. The accuracy of roundness and squareness in their machining tests was improved by factors of 3.7 and 7.0, respectively.

An investigation by Ihara et al. [53] using a Telescoping magnetic ball bar (DBB) resulted in a reduction in motion error of the platform to 25% of the uncalibrated values. They used Fourier transformation and included the length error of struts, position error of base & joint errors of the platform for optimizing the overall error of the system.

### 4.3 Constraint Approach

The constraint calibration approach is less popular due to the limitation of its practical application. This method is implemented by using some mechanical constraints on a passive joint of the PKM [26]. The already existing sensors in the constraint joints act as the measuring instruments; external sensors or measuring equipment are not normally used. The applied constraint causes a reduction in the workspace of the manipulator and reduces the sensitivity of the kinematics parameters. Also, the force generated due to constraining the movement may distort the structure and impacts the accuracy of the calibration.

Ryu et al. 2001 [26] & Rauf et al. 2001 [54] used this approach to calibrate a hexapod "Hexa Slide Machine (HSM)" by constraining its leg/s. Ryu constrained one leg of the system and the system was forced to behave like a 5-DOF system instead of the regular 6-DOF system. There was no extra sensor used, the in-built existing leg sensor was used for taking the required measurements. Rauf constrained 3 legs and experimented with 3-DOF. In both cases, the final correction values are near zero. By constraining one leg, the initial values of the platform position $8.0e^{-3}$ m and orientation $1.4e^{-2}$ rad converge to $3.8e^{-16}$ m & $1.7e^{-15}$ radians, respectively. And for 3-DOF measurements, initial values of position $8.0e^{-3}$ m and orientation $1.4e^{-2}$ radians changed to $2.4e^{-10}$ m & $1.9e^{-9}$ radians, respectively. The advantage of these methods is that the locking device can be universal and need not have to be specific for a particular system.

### 4.4 Auto or Self Approach

Self-calibration is one of the ways to calibrate Stewart platform-based 6-DOF machines. This method requires adding some redundant sensors to the passive joints. This is not an easy task and adds complexity to the design and manufacturing of PKMs. Moreover, adding redundant sensors can make system development more expensive. The auto or self-calibration method may limit the workspace for calibration. Some research studies had been done with this method.

In one of those studies, Chiu et al. [55] used a cylindrical gauge block and a commercial trigger probe to do the auto-calibration of the PKM. They made use of the non-linear least square method in their algorithm. The advantage of their method is that the instruments are standard and commercialized which makes them easily available and less expensive. Multiple PKMs were used to validate their method and the results showed that different levels of accuracies were achieved. Similarly, in another auto-calibration process, Zhuang et al. [17] used a Coordinate Measuring Machine (CMM) with their robot FAU Stewart Platform. Here also the position and orientation of the platform were used to calibrate the system error. The use of the Levenberg-Marquart algorithm gave them an error reduction by 50%. The highlights from their research are that some extra sensors were needed to be installed in some of the joints to gather calibration data and any error of the end-effector attached to the platform needed to have a separate calibration. The research done by Patel et al. [9] with Ball-Bar for calibration of their platform poses increased error in some cases and for around 90% of cases, they observed accuracy improvement from 50% to 100%. They used the least square method for their calibration algorithm. The advantage of their method is that the extra leg consisting of Ball-Bar can remain with the system during the actual machine operation to allow online calibration and the extra leg is easily mountable and un-mountable.

The details are presented in Table 5. It should be noted that for each case, platform pose error was considered.

**Table 5: Assessment of auto or self-calibration approaches**

| Reference | Robot Name | Instruments used | Methods | Performance | Important points |
|---|---|---|---|---|---|
| Chiu et al. 2003 [55] | - | A cylindrical gauge block and a commercial trigger probe | Nonlinear least squares | Multiple PKMs are used to validate the calibration process and different accuracy levels have been obtained. | The instruments used here are standardized and commercialized. The method is comparatively compact and economical. |
| Zhuang et al. 2000 [17] | FAU Stewart Platform | CMM | Levenberg-Marquardt algorithm | The average error was reduced by more than 50 percent. | The end-effector required separate calibration since it is not part of the closed-loop kinematic chains. It requires redundant sensors that need to be installed at some of the joints of the machine tool. |
| Patel et al. 2000 [9] | - | Ball-bar | Least square minimization | Though in some cases the error increased, for 90% of cases improvement observed between 50% to 100%. | The extra legs can be mounted or unmounted easily. When left with the machine in certain situations, it enables online calibration. |

## 5 DISCUSSION AND COMPARISON

The goal of each calibration method is to make the hexapod machines accurate, improve precision and obtain correct results at the platform pose during their operations. As mentioned earlier, this pose accuracy is dependent on several mechanical and surrounding factors like temperature and load being experienced by the system. The ideal calibration would consider all these factors for any working condition of the machine but achieving that is not only expensive and time-consuming but also potentially unnecessary depending on the application conditions. In the experiments where only the leg lengths have been considered as the sources of error for the accuracy of the platform pose, it should be noted that the motion of the platform is dependent on all the joints which are moving to generate the motion. So, while calibration may include only the leg length error, it also indirectly includes the error contributed by the joints. Even if all joints are not equipped with individual sensors, their error factors are indirectly accounted for in the calibration process. In general, when the platform pose accuracy is of primary interest, the calibration of it indirectly considers all the error factors existing in the hexapod system but adding appropriate compensation for each error factor becomes difficult unless they are correctly identified and accounted for in the calculations.

### 5.1 Error Identification

From the tables above, it is seen that the level of improvements obtained are varied, with cases showing an increase in the error values. The error factors considered also are different in each study. In most cases, the platform pose error due to the leg length errors remained common and was considered as the primary error contributor in the whole system. Also, considering the error of each leg separately and equally impacts the overall accuracy of the PKM. In all these calibration processes, use of external instruments is the most common practice. External instruments such as Double Ball Bar (DBB), laser interferometer, biaxial inclinometer, laser tracker, telescopic magnetic ball bar, etc. were used. A separate study may be useful to ascertain the effectiveness of each of these instruments for the calibration of a hexapod by inverse kinematics.

Optical calibration methods also gained importance in recent times. CCD camera, omnidirectional camera, digital video camera, and other image capturing devices were used to achieve an accuracy level in the range of 1.0cm to 0.1mm. The advantage of using optical methods is that the PKM system does not need modifications to accommodate the measurement equipment for calibration, also it is non-invasive and automatically records the event for future reference. In these cases, the quality of the optical systems and the associated analysis system configuration have a major contribution to the final accuracy and precision attained.

### 5.2 Algorithms

Among the several different algorithms used for the calibration purpose, the least-square method based on the Levenberg-Marquardt algorithm was used for both optical and non-optical calibration. The genetic algorithm had been used for a couple of studies, but they appeared to have high running time to reach an optimized level and were not sensitive to measurement noises. Constrained optimization method and Tabu search, Algebraic Elimination, non-linear least square methods were used and all of them resulted in various levels of calibration accuracy. The use of quaternions has improved the calculation efficiency, though the practical application of hexapod accuracy level achieved by this technique has yet to be evaluated. Different algorithms also yielded different results on the same system for the calibrations done by Daney et al. Therefore, the selection of calibration methods and algorithms plays an important role in obtaining the desired accuracy level for the PKM system and application. In all cases, the final platform pose is the guiding parameter to evaluate calibration outcome.

### 5.3 External factors

There are some inherent dimensional errors in the hexapod structure due to the dimensional tolerances of each component used in the fabrication. The cumulative effects of all these tolerances play a significant role in platform accuracy. Apart from these errors, other factors, hysteresis for instance, vary with the change of the operational characteristics. Normally, the effect of temperature is expected to be minimal for the PKM system unless the system experiences large temperature variations during its operation. From a practical point of view, such robot systems operate in a controlled environment unless they are deployed in special applications like large field telescope mounting. In these applications where exposure to fluctuating weather conditions is unavoidable, proper calibration factors for thermal deviation must be included.

Likewise, the elastic deformation error factor is not dominant in all cases. If the hexapod platform is subjected to high loads relative to its structure, a factor for elastic deflection is essential. Hexapod platforms reviewed in this paper can carry loads up to 2000kg. As such, the structure can undergo a substantial amount of elastic stress and that may lead to a notable amount of deflection to impact the accuracy of the robot. In these types of cases, the factor of elastic load cannot be ignored. There are potential research opportunities for evaluating the impact of load conditions on the PKM systems and to add suitable compensation factors for further improving the platform pose accuracy.

## 6    CONCLUSIONS

Hexapods are used for precise, complex, and repeatable operations in a variety of applications. Small errors in the system can lead to a serious impact on the resulting motion. Therefore, calibration is a critical activity for any hexapod machine to be reliable in a given application. There are several sources of errors that negatively impact the accuracy of the hexapod. Depending on the application field and the parameters taken into consideration, choosing a proper calibration method will significantly improve the accuracy. This paper attempts to provide an overview of those methods and draw an outline of the current state of the art in this field and help other researchers to take an appropriate note of the desired methods.

Several different calibration methods have been explored by the researchers, and they show the differing levels of accuracy and deployment complexities. The focus of this review is to identify the methods used to improve the positional accuracy of the hexapod platform or end-effector. Instruments and techniques used for incorporating compensation for the error generated due to different inaccuracies are discussed. Three types of calibration strategies are identified for physical systems: external, constraint, and auto- or self-calibration. Double ball bar laser trackers, and optical devices are some of the most commonly used instruments for collecting positional data. As for the types of errors considered, majority of the studies used platform pose or leg length. In addition to studies on physical systems, simulation-based studies are also discussed.

Research reviewed in this paper focused on calibrating the pose of the platform under no-load condition; however, in actual applications, these hexapods may be subjected to heavy working loads on the platform. The working load propagates directly to the hexapod structure. Based on the structural rigidity, the load causes elastic deformation which can affect the operational accuracy. Therefore, the response behavior and platform accuracy of hexapods under the influence of working loads remains a subject for further studies.

## REFERENCES

[1]     Harib K. H., Sharif Ullah A. M. M., and Hammami A., 2007, "A hexapod-based machine tool with hybrid structure: Kinematic analysis and trajectory planning,"

*Int. J. Mach. Tools Manuf.*, **47**(9), pp. 1426–1432.

[2] Shi H., She Y., and Duan X., 2015, "Modeling and measurement algorithm of hexapod platform sensor using inverse kinematics," *2015 Aust. Control Conf. AUCC 2015*, pp. 331–335.

[3] Zoran PANDILOV, and Rall K., 2012, "Parallel Kinematics Machine Tools : Overview - From History To the Future," *Mech. Eng. Sci. J.*, **25**(1), pp. 3–20.

[4] Jouini M., Sassi M., Sellami A., and Amara N., 2013, "Modeling and control for a 6-DOF platform manipulator," *2013 Int. Conf. Electr. Eng. Softw. Appl. ICEESA 2013*.

[5] Gough V. E., 1957, "Contribution to discussion to papers on research in automobile stability and control and in tire performance," *Proc. Auto Div. Inst. Mech. Eng.*, **171**, pp. 392–397.

[6] Stewart D., 1965, "A Platform with Six Degrees of Freedom," *Proc. Inst. Mech. Eng. , Part I J. Syst. Control Eng.*

[7] Wu J. F., Zhang R., Wang R. H., and Yao Y. X., 2014, "A systematic optimization approach for the calibration of parallel kinematics machine tools by a laser tracker," *Int. J. Mach. Tools Manuf.*, **86**, pp. 1–11.

[8] Mazare M., Taghizadeh M., and Rasool Najafi M., 2017, "Kinematic analysis and design of a 3-DOF translational parallel robot," *Int. J. Autom. Comput.*, **14**(4), pp. 432–441.

[9] Patel A. J., and Ehmann K. F., 2000, "Calibration of a hexapod machine tool using a redundant leg," *Int. J. Mach. Tools Manuf.*, **40**(4), pp. 489–512.

[10] Patel Y. D., and George P. M., 2012, "Parallel Manipulators Applications—A Survey," *Mod. Mech. Eng.*, **02**(03), pp. 57–64.

[11] Physik Instrumente (PI) GmbH & Co KG., 2018, "Hexapod Parallel Robots Automate Highly Precise Production Processes," *AZoNano*, pp. 1–17.

[12] Majarena A. C., Santolaria J., Samper D., and Aguilar J. J., 2010, "An overview of kinematic and calibration models using internal/external sensors or constraints to improve the behavior of spatial parallel mechanisms," *Sensors (Switzerland)*, **10**(11), pp. 10256–10297.

[13] Merlet J., 2002, "Still a long way to go to the road for parallel mechanisms," *ASME 2002 DETC Conf. Montréal, Canada*, (May).

[14] Daney D., Papegay Y., and Neumaier A., 2004, "Interval methods for certification of the kinematic calibration of parallel robots," *Proc. - IEEE Int. Conf. Robot. Autom.*, **2004**(2), pp. 1913–1918.

[15] Jáuregui J. C., Hernández E. E., Ceccarelli M., López-Cajún C., and García A., 2013, "Kinematic calibration of precise 6-DOF Stewart platform-type positioning systems for radio telescope applications," *Front. Mech. Eng.*, **8**(3), pp. 252–260.

[16] Ziegert J. C., Jokiel B., and Huang C.-C., 1999, "Calibration and Self-Calibration of Hexapod Machine Tools," *Parallel Kinematic Mach. © Springer-Verlag London Ltd. 1999*, pp. 205–216.

[17] Zhuang H., Liu L., and Masory O., 2000, "Autonomous calibration of hexapod machine tools," *J. Manuf. Sci. Eng. Trans. ASME*, **122**(1), pp. 140–148.

[18] Nubiola A., and Bonev I. A., 2014, "Absolute robot calibration with a single telescoping ballbar," *Precis. Eng.*, **38**(3), pp. 472–480.

[19] Szep C., Stan S. D., Csibi V., Manic M., and Bălan R., 2009, "Kinematics, workspace, design and accuracy analysis of RPRPR medical parallel robot," *Proc. - 2009 2nd Conf. Hum. Syst. Interact. HSI '09*, pp. 75–80.

[20] Bleicher F., Puschitz F., and Theiner A., 2006, "Laser based measurment system for calibrating machine tools in 6 dof," *Ann. DAAAM Proc. Int. DAAAM Symp.*, pp. 39–40.

[21] Ramesh R., Mannan M. ., and Poo A. ., 2000, "Error compensation in machine tools — a review," *Int. J. Mach. Tools Manuf.*, **40**(9), pp. 1257–1284.

[22] Szatmári S., 1999, "Geometrical errors of parallel robots," *Period. Polytech. Mech. Eng.*, **43**(2), pp. 155–162.

[23] Daney D., 2002, "Optimal measurement configurations for Gough platform calibration," *Proc. - IEEE Int. Conf. Robot. Autom.*, **1**(May), pp. 147–152.

[24] Soons J. A., 1997, "Error analysis of a hexapod machine tool," *Trans. Eng. Sci.*, **16**, pp. 347–358.

[25] Traslosheros A., Sebastián J. M., Torrijos J., Carelli R., and Castillo E., 2013, "An inexpensive method for kinematic calibration of a parallel robot by using one hand-held camera as main sensor," *Sensors (Switzerland)*, **13**(8), pp. 9941–9965.

[26] Ryu J., and Rauf A., 2001, "A new method for fully autonomous calibration of parallel manipulators using a constraint link," *IEEE/ASME Int. Conf. Adv. Intell. Mechatronics, AIM*, **1**(July), pp. 141–146.

[27] Traslosheros A., Sebastián J. M., Castillo E., Roberti F., and Carelli R., 2011, "A method for kinematic calibration of a parallel robot by using one camera in hand and a spherical object," *IEEE 15th Int. Conf. Adv. Robot. New Boundaries Robot. ICAR 2011*, pp. 75–81.

[28] Ren X. D., Feng Z. R., and Su C. P., 2009, "A new calibration method for parallel kinematics machine tools using orientation constraint," *Int. J. Mach. Tools Manuf.*, **49**(9), pp. 708–721.

[29] Huang T., Wang J., Chetwynd D. G., and Whitehouse D. J., 2003, "Identifiability of geometric parameters of 6-DOF PKM systems using a minimum set of pose error data," *Proc. - IEEE Int. Conf. Robot. Autom.*, **2**(1), pp. 1863–1868.

[30] Zou H., and Notash L., 2001, "Discussions on the camera-aided calibration of parallel manipulators," *Proc. 2001 CCToMM Symp. Mech. Mach. Mechatronic, Saint-Hubert*, pp. 3–4.

[31] Olarra A., Axinte D., and Kortaberria G., 2018, "Geometrical calibration and uncertainty estimation methodology for a novel self-propelled miniature

robotic machine tool," *Robot. Comput. Integr. Manuf.*, **49**(January 2017), pp. 204–214.

[32] Meng Y., and Zhuang H., 2007, "Autonomous robot calibration using vision technology," *Robot. Comput. Integr. Manuf.*, **23**(4), pp. 436–446.

[33] Zhuang H., Yan J., and Masory O., 1998, "Calibration of Stewart platforms and other parallel manipulators by minimizing inverse kinematic residuals," *J. Robot. Syst.*, **15**(7), pp. 395–405.

[34] Santolaria J., and Ginés M., 2013, "Uncertainty estimation in robot kinematic calibration," *Robot. Comput. Integr. Manuf.*, **29**(2), pp. 370–384.

[35] Everett L. J., Driels M., and Mooring B. W., 1987, "Kinematic Modelling for Robot Calibration.," *IEEE*, pp. 183–189.

[36] Vischer P., and Clavel R., 1998, "Kinematic calibration of the parallel Delta robot," *Robotica*, **16**(2), pp. 207–218.

[37] Yang X. L., Wu H. T., Li Y., Kang S. Z., and Chen B., 2019, "Computationally Efficient Inverse Dynamics of a Class of Six-DOF Parallel Robots: Dual Quaternion Approach," *J. Intell. Robot. Syst. Theory Appl.*, **94**(1), pp. 101–113.

[38] Agheli M., and Nategh M., 2009, "Identifying the Kinematic Parameters of Hexapod Machine Tool.," *World Acad. Sci. Eng. Technol.*, **52**, pp. 380–385.

[39] Daney D., and Emiris I., 2001, "Variable elimination for reliable parallel robot calibration," *2nd Work. Comput. Kinemat. - CK'2001 12 p*, p. 13.

[40] Daney D., and Emiris I. Z., 2004, "Algebraic Elimination for Parallel Robot Calibration," *Proc. 11 World Congr. Mech. Mach. Sci. Tianjin China*.

[41] Wang S. M., and Ehmann K. F., 2002, "Error model and accuracy analysis of a six-DOF Stewart Platform," *J. Manuf. Sci. Eng. Trans. ASME*, **124**(2), pp. 286–295.

[42] Mahmoodi A., Sayadi A., and Menhaj M. B., 2014, "Solution of forward kinematics in Stewart platform using six rotary sensors on joints of three legs," *Adv. Robot.*, **28**(1), pp. 27–37.

[43] Nategh M. J., and Agheli M. M., 2009, "A total solution to kinematic calibration of hexapod machine tools with a minimum number of measurement configurations and superior accuracies," *Int. J. Mach. Tools Manuf.*, **49**(15), pp. 1155–1164.

[44] Großmann K., Kauschinger B., and Szatmári S., 2008, "Kinematic calibration of a hexapod of simple design," *Prod. Eng.*, **2**(3), pp. 317–325.

[45] Liu Y., Liang B., Li C., Xue L., Hu S., and Jiang Y., 2007, "Calibration of a Steward parallel robot using genetic algorithm," *Proc. 2007 IEEE Int. Conf. Mechatronics Autom. ICMA 2007*, pp. 2495–2500.

[46] Ting Y., Jar H. C., and Li C. C., 2007, "Measurement and calibration for Stewart micromanipulation system," *Precis. Eng.*, **31**(3), pp. 226–233.

[47] Daney D., Andreff N., Chabert G., and Papegay Y., 2006, "Interval method for calibration of parallel robots: Vision-based experiments," *Mech. Mach. Theory*, **41**(8), pp. 929–944.

[48] Dallej T., Hadj-Abdelkader H., Andreff N., and Martinet P., 2006, "Kinematic Calibration of a Gough-Stewart platform using an omnidirectional camera," *IEEE Int. Conf. Intell. Robot. Syst.*, pp. 4666–4671.

[49] Daney D., Papegay Y., and Madeline B., 2005, "Choosing measurement poses for robot calibration with the local convergence method and Tabu search," *Int. J. Rob. Res.*, **24**(6), pp. 501–518.

[50] Gao M., Li T., and Yin W., 2003, "Calibration method and experiment of Stewart platform using a laser tracker," *Proc. IEEE Int. Conf. Syst. Man Cybern.*, **3**, pp. 2797–2802.

[51] Renaud P., Andreff N., Dhome M., and Martinet P., 2002, "Experimental evaluation of a vision-based measuring device for parallel machine-tool calibration," *IEEE Int. Conf. Intell. Robot. Syst.*, **2**(October), pp. 1868–1873.

[52] Week M., and Staimer D., 2002, "Accuracy issues of parallel kinematic machine tools," *Proc. Inst. Mech. Eng. Part K J. Multi-body Dyn.*, **216**(1), pp. 51–57.

[53] Ihara Y., Ishida T., Kakino Y., Li Z., Matsushita T., and Nakagawa M., 2000, "Kinematic calibration of a hexapod machine tool by using circular test," *Proc. 2000 Japan•USA Flex. Autom. Conf. , Ann Arbor, Michigan*, (January), pp. 1–4.

[54] Rauf A., and Ryu J., 2001, "Fully autonomous calibration of parallel manipulators by imposing position constraint," *Proc. - IEEE Int. Conf. Robot. Autom.*, **3**, pp. 2389–2394.

[55] Chiu Y. J., and Perng M. H., 2003, "Self-calibration of a general hexapod manipulator using cylinder constraints," *Int. J. Mach. Tools Manuf.*, **43**(10), pp. 1051–1066.